\documentclass[runningheads]{llncs}

\usepackage{eccv}

\usepackage{eccvabbrv}

\usepackage{graphicx}
\usepackage{amsmath,amssymb,amsfonts,dsfont}
\usepackage{booktabs}

\usepackage[accsupp]{axessibility}  % Improves PDF readability for those with disabilities.

\usepackage{hyperref}

\usepackage{orcidlink}

\begin{document}

\title{Instance-wise Uncertainty for Class Imbalance in Semantic Segmentation} 


\author{Luís Almeida \inst{1} \orcidlink{0009-0004-0705-1801} \and
Inês Dutra \inst{1} \orcidlink{0000-0002-3578-7769} \and
Francesco Renna \inst{1} \orcidlink{0000-0002-8243-8350}}

\authorrunning{L. Almeida, I. Dutra, et. al}

\institute{Faculdade de Ciências da Universidade do Porto \\
\url{https://www.up.pt/fcup/en/}}

\maketitle

\begin{abstract}
Semantic segmentation is a fundamental computer vision task with a vast number of applications. State of the art methods increasingly rely on deep learning models, known to incorrectly estimate uncertainty and being overconfident in predictions, especially in data not seen during training. This is particularly problematic in semantic segmentation due to inherent class imbalance. Popular uncertainty quantification approaches are task-agnostic and fail to leverage spatial pixel correlations in uncertainty estimates, crucial in this task. In this work, a novel training methodology specifically designed for semantic segmentation is presented and tested in datasets of road scene images. Training samples are weighted by the uncertainty of each class instance estimated by an ensemble. This is shown to increase performance on minority classes, boost model generalization and robustness to domain-shift when compared to using the inverse of class proportions or no class weights at all. This method addresses the challenges of class imbalance and uncertainty estimation in semantic segmentation, potentially enhancing model performance and reliability across various applications.
\end{abstract}

\section{Introduction}
\label{sec:intro}

Semantic segmentation stands as a fundamental task in the computer vision domain, finding widespread adoption in areas such as autonomous driving \cite{8317714,18631,9913352}, biomedical imaging \cite{article,Kar2021,9330594} and more. This task involves classifying each pixel in an image into predefined categories, enabling detailed scene understanding crucial for various applications. State of the art semantic segmentation approaches predominantly employ deep earning solutions due to their exceptional performance. However, despite remarkable advancements in deep learning algorithms, the reliability and robustness of deployed AI systems remain critical concerns. These models often exhibit overconfidence when encountering unfamiliar data domains and lack robust mechanisms for estimating prediction uncertainty \cite{DBLP:journals/corr/GuoPSW17}. Addressing the challenge of uncertainty modeling and quantification in these models is a pivotal step in advancing the field.

One of the major challenges in semantic segmentation tasks is class imbalance. This problem arises because deep learning models tend to overfit to the best-represented classes, leading to suboptimal performance on worst-represented classes \cite{BRESSAN2022102690}. Class imbalance is particularly problematic in scenarios where the distribution of classes is inherently skewed, such as in biomedical imaging, where certain pathologies are rare, or in autonomous driving, where certain objects may be less frequently encountered.

The primary solution for mitigating class imbalance involves weighting the classes in the loss function according to the proportion of pixels of each class present in the data. By assigning higher weights to worse-represented classes, the model is encouraged to focus more on them during training, thereby improving its performance. However, this is typically not enough to significantly improve model performance on minority classes, and may even induce it to produce miscalibrated confidence scores \cite{caplin2022calibratingclassweightsmodeling}.

In this work, a novel training method for semantic segmentation is proposed. It leverages predictive uncertainty at the instance level to weight the loss function, thus making the model focus more on pixels of harder classes. Through diverse experiments, it is shown that this novel method not only increases the performance on minority classes but is also capable to provide good uncertainty estimates and greatly increase robustness to domain shift. It was also found that models adopting this training methodology significantly performed better on minority classes than models using the inverse of class proportions as class weights in the loss function. This makes the usage of instance-wise uncertainty as weights in the loss function a great option for imbalanced datasets.

\section{Related Work}

\subsection{Background}

In the context of Machine Learning, uncertainty is usually decomposed into two different components according to its source, referred to as aleatoric uncertainty and epistemic uncertainty \cite{KIUREGHIAN2009105}. Aleatoric uncertainty (AU) encompasses the uncertainty that stems from the data, mainly due to noise and the complexity of the process that generated it \cite{malinin2019uncertainty}. Epistemic uncertainty (EU) refers to the uncertainty that arises from ignorance about the best model, i.e., from unawareness about the best model assumptions and uncertainty about its parameters. The predictive uncertainty for an input sample $x$ is the sum of its epistemic uncertainty and aleatoric uncertainty. The most prominent way of formalizing these concepts is by utilizing a Bayesian framework \cite{pmlr-v37-blundell15}. Let $Y$ be a vector of class probabilities for $x$, $D$ a dataset and $w$ a set of model weights. The predictive distribution $P(Y|x,D)$ is given by

\begin{equation}
    \label{eq:preddist}
    P(Y|x,D) = \mathds{E}_{P(w|D)}[P(Y|x,w)]
\end{equation}

The predictive uncertainty for a given input $x$ is then given by the entropy $H$ of the predictive distribution of $x$, i.e.,

\begin{equation}
    \label{eq:predunc}
    \text{Predictive Uncertainty} = H[\mathds{E}_{P(w|D)}[P(Y|x,w)]]
\end{equation}

Aleatoric uncertainty is often defined as the expected entropy of $P(Y|x,w)$ and epistemic uncertainty by the mutual information (MI) between $Y$ and the posterior over the model parameters \cite{smith2018understandingmeasuresuncertaintyadversarial}.
\begin{align}
    \text{AU} &= \mathds{E}_{P(w|D)}[H[P(Y|x,w)]] & \quad & \text{EU} = MI(Y,w | D,x)
\end{align}

It is worth noting that recently, some doubts regarding how appropriate these definitions are\cite{wimmer2023quantifying}, how well AU and EU can be separated \cite{kahl2024values} and how necessary this decomposition is \cite{KIUREGHIAN2009105} have been raised. Throughout this work, AU and EU will not be distinguished and only the concept of predictive uncertainty will be used.

\subsection{Uncertainty Quantification in Semantic Segmentation}

The problem of uncertainty quantification (UQ) has been extensively studied across several domains and tasks, such as image classification, image semantic segmentation and natural language processing. Popular methods in the literature such as Monte-Carlo (MC) Dropout \cite{pmlr-v48-gal16} and its variants \cite{Mobiny2021,fastdropouttraining}, Deep Ensembles \cite{NIPS2017_9ef2ed4b} and Variational Inference (VI) \cite{10.1007/s10462-023-10562-9} are task-agnostic and have all been used for quantifying uncertainty in semantic segmentation tasks. All these methods implicitly approximate (\ref{eq:preddist}) and use a dispersion metric (such as variance or entropy) to measure uncertainty.

MC Dropout uses the dropout \cite{JMLR:v15:srivastava14a} technique to effectively "turn off" neurons and to obtain a different model prediction in each forward pass, with uncertainty being the variance of the confidence scores. This can be viewed as sampling different sets of model architectures to approximate $P(w|D)$. However, since dropout is applied randomly, it may yield architectures with high loss values in each sample and lead to poor uncertainty estimates \cite{DBLP:journals/corr/abs-2110-04286}.

Ensembles were first proposed to quantify uncertainty by randomly initializing the weights of $M$ different $f_m$ models (with the same architecture) and independently training them on the same dataset. The confidence scores of these models are averaged into a single predictive distribution and the uncertainty is given by its entropy, i.e., $H[\frac{1}{M}\sum_{m = 1}^M f_m(x)]$ . Thus, ensembles rely on random initialization of model weights to foster predictive diversity and to get low loss samples from $P(w|D)$ by training these models. Nonetheless, random initialization of model weights does not allow for the usage of transfer learning techniques and has been found to be lackluster in achieving predictive diversity. To solve this problem, approaches such as using different architectures in the ensemble \cite{9956231} have been proposed.

VI is another method that has been used to approximate the posterior $P(w|D)$ \cite{10.1007/s10462-023-10562-9}. Using a family of parametric distributions $Q(w)$, VI finds the parameters that minimize the difference between $Q(w)$ and the posterior. This difference between distributions is measured by the Kullback-Leibler Divergence in (\ref{eq:kldiverg}):

\begin{equation}
    \label{eq:kldiverg}
    KL(Q||P) = \mathds{E}_Q \left[ \log \frac{Q(w)}{P(w|D)}\right]
\end{equation}
Since this equation explicitly uses the posterior, it cannot be directly minimized. Thus, the Evidence Lower Bound (\ref{eq:elbo}) is maximized instead, since this is equivalent to minimizing the KL divergence up to an added constant.

\begin{equation}
    \label{eq:elbo}
    ELBO = \mathds{E}_Q \left[ \log \frac{P(y|x,w)}{Q(w)}\right]
\end{equation}

The KL divergence can then be obtained by $KL(Q||P) = - ELBO + \log P(y|x)$ \cite{10.1007/s10462-023-10562-9}.
One pioneering work that leverages VI is Bayes By Backprop \cite{pmlr-v37-blundell15}. By using the Gaussian reparameterization trick \cite{Opper2009}, it is possible to approximate the ELBO as
\begin{equation}
    F(D,\theta) \simeq \sum_{i=1}^{n}\log Q(w_i|\theta)-\log P(w_i) - \log P(D|w_i)
\end{equation}
where $\theta$ represents the parameters of the variational posterior, which is chosen to be a diagonal Gaussian distribution with mean $\mu$ and standard deviation $\rho$. For the prior distribution $P(w)$, a scale mixture of two Gaussians is used. The transform that yields a sample of weights $w$ is defined as $w = \mu + \log(1+\exp(\rho)) \cdot \epsilon$, where $\epsilon \sim \mathcal{N}(0,I)$. In each optimization step, a sample of weights is drawn according to this transform and these are then used in the model to compute the gradients with the usual backpropagation algorithm. Parameters $\mu$ and $\rho$ are then updated according to these gradients. Unfortunately, VI suffers from problems such as determining an appropriate variational family of distributions and the bias which is introduced by said family, as the true posterior $P(w|D)$ is often more complex than $Q(w)$. For semantic segmentation, it has also been pointed that works using variational inference relying on sampling for the ELBO gradients estimation is not viable in semantic segmentation due to its high-dimensionality of inputs and outputs\cite{Carvalho_2020_CVPR}.  

UQ methodologies that do not try to approximate the posterior also exist. Deep Deterministic Uncertainty \cite{DBLP:journals/corr/abs-2102-11582} is a single model approach that models each class in feature space using Gaussian discriminant analysis, which enables EU and AU estimates to be obtained. A similar method leveraging Quantum Information Potential Field to decompose the probability density function of the data in feature space to a reproducing kernel hilbert space has also been introduced in \cite{singh2022quantifyingmodeluncertaintysemantic}. However, it has been argued that single model UQ provides poorer uncertainty estimates than the previously presented approaches \cite{postels2022practicalitydeterministicepistemicuncertainty}.

Most of these works propose task-agnostic UQ methods which do not leverage existing pixel correlations that are crucial in semantic segmentation. Furthermore, works that propose the usage of uncertainty values to improve model performance have limitations. Closely related to this work,\cite{landgraf2023uceuncertaintyawarecrossentropysemantic} computes pixel-wise uncertainty masks using MC Dropout and uses them to weight the cross entropy loss function. However, MC Dropout requires multiple forward passes to compute the uncertainty of an input, which significantly slows down training. Additionally, by using pixel-wise uncertainty as weights for each pixel in the loss function, one fails to consider the uncertainty of the neighborhood pixels.

\section{Semantic Segmentation with Instance Uncertainty}
In this section, a novel method for semantic segmentation using instance-based uncertainty estimates is presented. Semantic segmentation can be viewed as a classification problem where the goal is to classify each pixel of an image as belonging to one of $C$ classes. Thus, the output of a semantic segmentation method is a segmentation mask where each pixel is assigned a class. Since the classes in semantic segmentation problems are usually imbalanced due to the inherent difference of the number of pixels in each of them, models have more difficulty learning features of the minority classes. The predictive uncertainty of the model associated with each pixel can be viewed as a measure of how hard it is to classify, thus can be used as a weight in the loss function to make the model focus more on these pixels. A simple method to improve the performance of semantic segmentation models in less well represented classes is then achieved by solving the following problems:

\begin{enumerate}
    \item Quantifying the uncertainty for each pixel of each training sample.\\

    \item Computing the uncertainty for each instance of a class in each training sample.\\
    
    \item Using the estimated uncertainties to weight the training sample.\\
\end{enumerate}

\begin{figure}
    \centering
    \includegraphics[height=8cm,width=10cm]{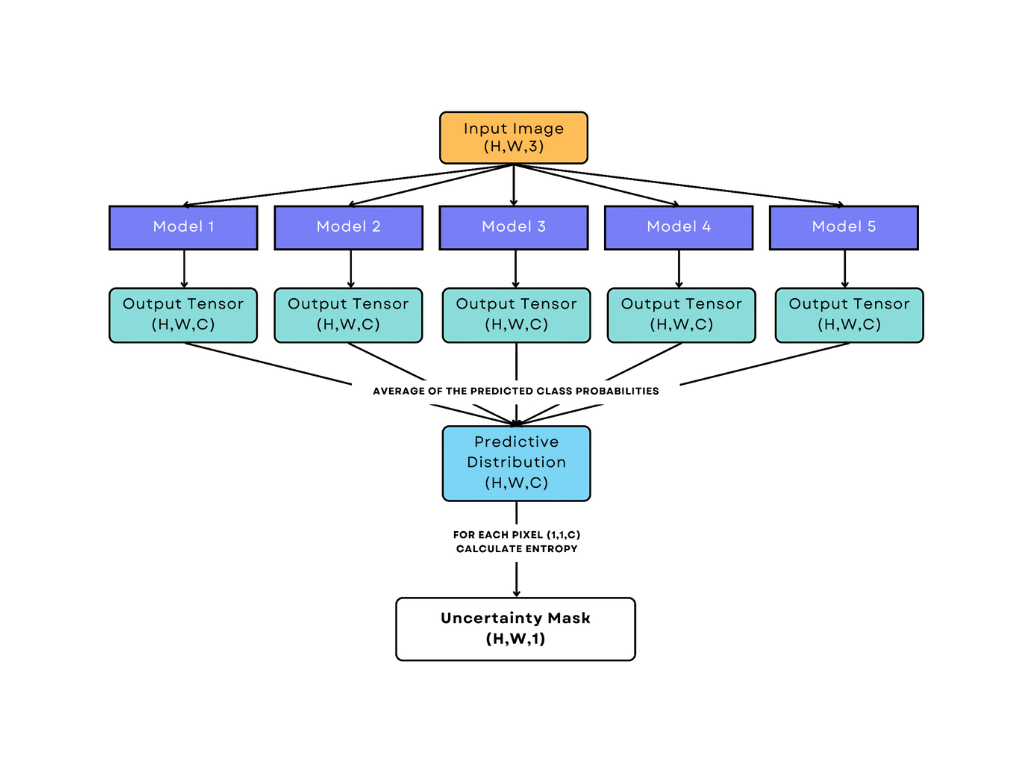}
    \caption{Computing the pixel-wise uncertainty mask of an input image using an ensemble of five models.}
    \label{fig:ensembles}
\end{figure}

To compute the pixel-wise uncertainty for each training sample, an UQ method can be adapted and used for semantic segmentation. Since ensembles have been known to achieve state of the art performance on UQ tasks, are simple to implement and can be run in parallel, this was the adopted approach to solve problem 1. Ensembles can generate pixel-wise uncertainty masks by combining the output confidence scores of each model, thus obtaining a predictive distribution for each pixel. The entropy of each of these distributions is then computed, resulting in pixel-wise uncertainty masks. An illustration of this procedure is given in Figure \ref{fig:ensembles}.

\begin{figure}[tb]
  \centering
  \begin{subfigure}[b]{0.4\textwidth}
    \includegraphics[width=\textwidth]{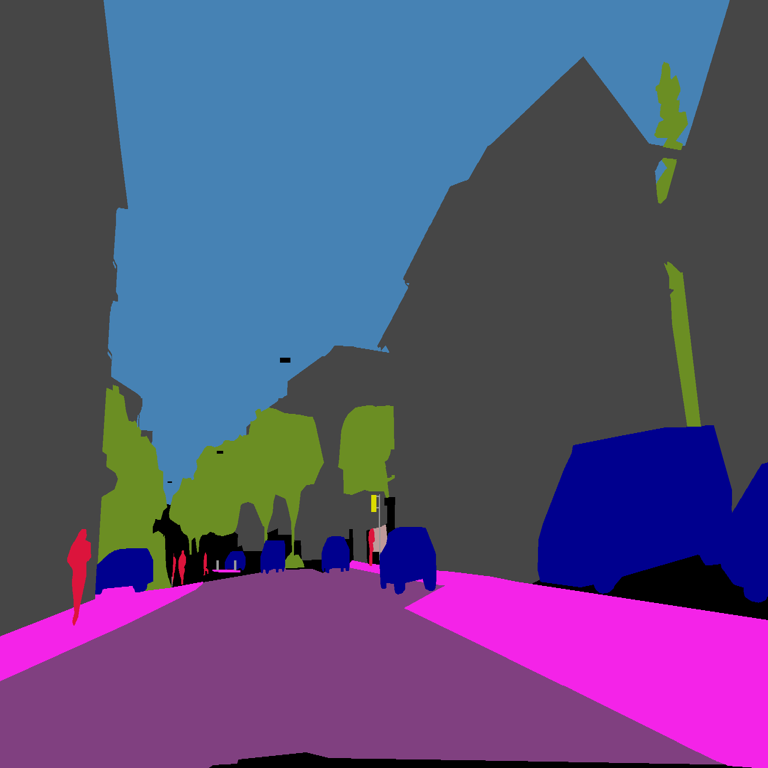}
    \caption{Segmentation mask of a training sample of the ACDC dataset. Here, each class is colored differently.}
    \label{fig:seg_mask}
  \end{subfigure}
  \hfill
  \begin{subfigure}[b]{0.4\textwidth}
    \includegraphics[width=\textwidth]{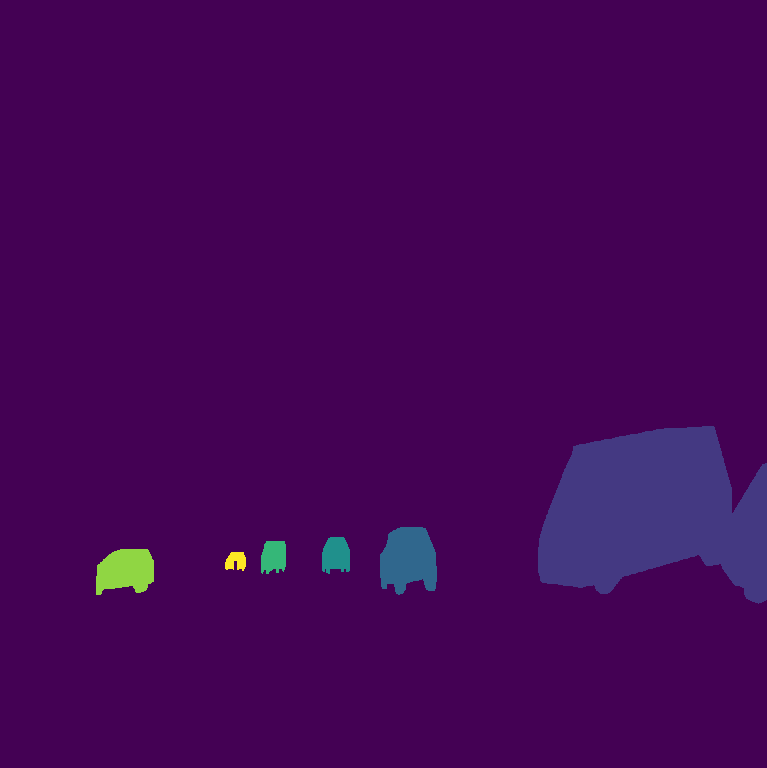}
    \caption{All instances of class car as computed by the described method. In this figure, each instance is colored differently.}
    \label{fig:instance_mask}
  \end{subfigure}
  \caption{Obtaining all instances of class car in a segmentation mask through the described strongly connected components methodology.}
  \label{fig:comparison}
\end{figure}

The purpose of using instance-wise uncertainty instead of settling for the pixel level is to incorporate the uncertainty information of the region where each pixel belongs in their weight for the loss function. Independent pixel uncertainty estimates may result in irregular weights, therefore hindering the training process. This can be circumvented by using the average uncertainty of the pixels in each region of the image, yielding smoother weights. However, choosing a meaningful neighborhood region to consider for each pixel is not obvious and can be done in multiple ways. One possible solution is to use the average uncertainty of the pixels in each instance. An instance of class $c$ is defined as a set of pixels that belong to $c$ such that there is a sequence of adjacent pixels in the set that connects each of its pixels. This is illustrated in Figures \ref{fig:seg_mask} and \ref{fig:instance_mask}.

For each pixel $p$, its instance uncertainty value is given by (\ref{eq:instance_unc}), where $I$ denotes an instance from the set $R$ of all instances in the input and PU denotes the pixel-wise predictive uncertainty. $\mathbf{1}_I(p)$ denotes the indicator function, that is, $\mathbf{1}_I(p) = 1$ if pixel $p$ belongs to instance $I$ and is zero otherwise.
\begin{equation}
\label{eq:instance_unc}
   \text{Instance Uncertainty}(p) = \sum_{I \in R} \mathbf{1}_I(p) \cdot \frac{\sum_{k \in I} \text{PU}(k)}{|I|}
\end{equation}

The problem of computing all instances for each class can then be solved by viewing the image as a connected graph and instances as strongly connected components. By determining all connected components in each input and their corresponding average uncertainty, an instance-wise uncertainty mask is obtained, thereby solving problem 2.

The most intuitive and direct way of using these instance-wise uncertainty masks to weight the loss function is by computing the Hadamard product between them and the cross entropy (CE) output tensor. However, since the uncertainty estimate for a pixel can be smaller than $1$, this can heavily decrease the weights of low uncertainty pixels in the loss function and hinder model performance. This can be mitigated by adding $1$ to each element in the uncertainty mask. Solving problem 3, the uncertainty weighted loss is given by (\ref{eq:loss}), where $\hat{y}$ and $y$ denote the tensor of predicted logits and the ground truth, respectively. $IU_{\hat{y}}$ denotes the computed instance uncertainty mask and $\odot$ denotes the Hadamard product. 

\begin{equation}
 \label{eq:loss}
     \mathcal{L}(\hat{y},y) = CE(\hat{y},y) \odot (1+IU_{\hat{y}})^2
\end{equation}

\section{Implementation}

The first step in building an ensemble is choosing the model architectures that integrate it. Considering a single architecture and randomly initializing model weights prohibits the usage of pre-trained model weights through transfer learning, which is known to enhance the performance of deep learning models. Additionally, due to the problems in achieving ensemble diversity through this method, three different backbone architectures (MobileNetV2 \cite{DBLP:journals/corr/HowardZCKWWAA17}, ResNet50 and ResNet101 \cite{resnet50}) of a DeepLabV3+ \cite{DBLP:journals/corr/abs-1802-02611} segmentation model were chosen. Although the original Deep Ensembles work suggests using five models, it is argued that reducing the number of models in the ensemble to three does not significantly impact performance \cite{ashukha2021pitfalls,9533330}, thus motivating the choice of ensembling only three different models. Differing at the stage where the uncertainty masks are computed, two approaches are considered:

\begin{enumerate}
    \item Train models $M_1,M_2,M_3$ independently on the training set, using the standard cross entropy loss. Build an ensemble using these models and compute the instance-wise uncertainty masks for each training sample. Train new models $M_1', M_2', M_3'$ by incorporating these uncertainty masks in their loss function, given by (\ref{eq:loss}). The advantage in this scenario is that by using good performing models, the uncertainty masks will be of higher quality. However, this adds the computational overhead of training an ensemble of three models solely to obtain uncertainty estimates.

    \item Build an ensemble using models $M_1,M_2,M_3$ and train them in parallel. For each training sample, obtain the instance-wise uncertainty mask and use it to compute each individual model loss function (\ref{eq:loss}). This way, the models directly compute uncertainty masks for training on the fly, without the need for a previous ensemble to generate them. The disadvantage is that models are weak predictors in the beginning of their training and can thus produce low quality uncertainty masks, resulting in bad weights for their loss function.  
\end{enumerate}

\begin{figure}[tb]
    \centering
    \includegraphics[width=12cm,height=10cm]{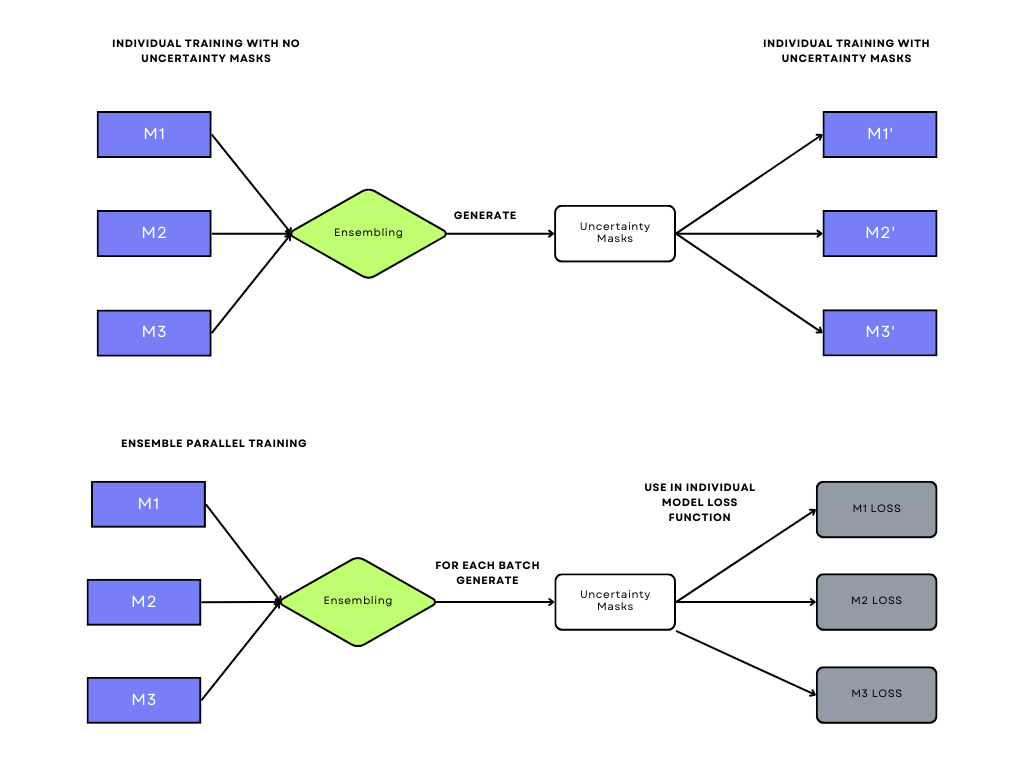}
    \caption{The considered approaches. In the first approach (top) the uncertainty masks are computed after the training of an ensemble. In the second approach (bottom) uncertainty masks are computed during the training process on the fly.}
    \label{fig:imp}
\end{figure}

Both of these approaches are illustrated in Figure \ref{fig:imp}.

\section{Evaluation Metrics}
For the semantic segmentation task, the Intersection-Over-Union (IoU) metric is the most popular to assess model performance. For a given class $c$, its IoU value is given by equation (\ref{eq:iou}). $TP_c$ denotes the number of class $c$ pixels that were correctly classified (True Positives), $FN_c$ denotes the number of class $c$ pixels that were classified as belonging to another class (False Negatives) and $FP_c$ denotes the number of pixels of other classes that were incorrectly classified as belonging to class $c$ (False Positives). The mean Intersection-over-Union (mIoU) is simply the average of each class IoU.

\begin{align}
\label{eq:iou}
    IoU(c) = \frac{TP_c}{TP_c + FN_c + FP_c} & \quad & mIoU = \sum_{c = 1}^C \frac{IoU(c)}{C}
\end{align}

Although several uncertainty metrics have been proposed, such as expected calibration error \cite{Naeini2015ObtainingWC} and the Brier score \cite{Brier1950VERIFICATIONOF}, accurately and consistently measuring uncertainty is still an open problem. Specific to semantic segmentation, the Patch Accuracy vs Patch Uncertainty metric \cite{DBLP:journals/corr/abs-1811-12709} (PAvPU) is used to evaluate the degree to which a model correctly estimates uncertainty. A patch is defined as a window of $w$ by $w$ pixels. Each patch in an image is grouped into four classes according to two criteria: Accuracy and Certainty. If the number of correctly classified pixels in a patch divided by its size is greater than some threshold (usually $0.5$), the patch is considered accurate and inaccurate otherwise. If the estimated mean uncertainty in a patch is lower than some threshold (the mean pixel uncertainty of the training set is recommended), the patch is considered certain and uncertain otherwise. Let $n_{ac},n_{ic},n_{au},n_{iu}$ be the number of patches that are accurate and certain, inaccurate and uncertain, accurate and uncertain and inaccurate and uncertain, respectively. The PAvPU is given by (\ref{eq:pavpu}).

\begin{equation}
\label{eq:pavpu}
    PAvPU = \frac{n_{ac}+n_{iu}}{n_{ac}+n_{iu}+n_{ai}+n_{ic}}
\end{equation}

% \noindent\begin{tabularx}{\textwidth}{@{}XX@{}}
% \begin{equation}
%     \label{eq:iou}
%     IoU(c) = \frac{TP_c}{TP_c + FN_c + FP_c}
% \end{equation} &
%   \begin{equation}
%   \label{eq:miou}
%   mIoU = \sum_{c = 1}^C \frac{IoU(c)}{C}
%   \end{equation}
% \end{tabularx}

\section{Experiments}

In this section, all of the performed experiments to test the proposed framework are reported. The CityScapes \cite{DBLP:journals/corr/CordtsORREBFRS16} and ACDC \cite{DBLP:journals/corr/abs-2104-13395} datasets were used as benchmarks due to their compatibility and extensive study in the literature. The CityScapes dataset contains 5000 images with high quality pixel level annotations of 19 different classes of road scenes. Of these, 2975 are training images, 500 for validation and 1525 for testing. The ACDC dataset was designed to be compatible with CityScapes, as it is comprised of 4006 road scene images of the same classes as CityScapes. However, unlike CityScapes, its images are captured in adverse weather conditions. The public dataset contains 1600 training images and 406 validation images. The test set is private. 

\subsection{Measuring the impact of instance-wise uncertainty in model performance}
\begin{figure}[tb]
    \centering
    \includegraphics[height=5cm,width=11cm]{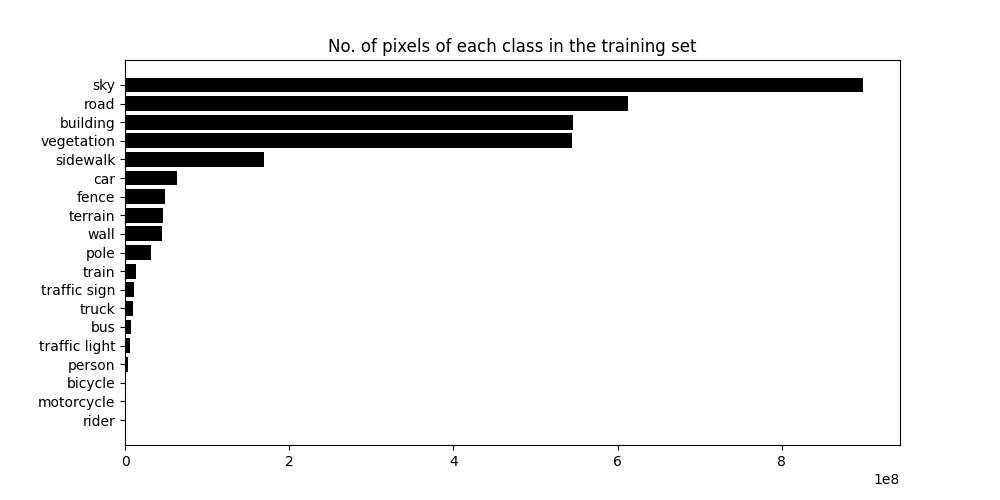}
    \caption{Histogram depicting the number of pixels belonging to each class in the ACDC training set.}
    \label{fig:pix_count}
\end{figure}

The goal of the first experiment was to understand how weighting the loss function using uncertainty estimates impacts model performance. For this reason, only a single dataset (ACDC) was used. Since ACDC's test set is private, all results come from the validation set. The first proposed method was tested by training three different DeepLabV3+ models with different backbone architectures, as mentioned in the preceding section. Instance-wise uncertainty masks were then generated using an ensemble of said models. Thirty different models for each distinct backbone architecture were trained, both with uncertainty masks and without uncertainty masks, resulting in a total of 180 total models trained. In Figure \ref{fig:pix_count}, the pixel distribution for each class is reported. The rider, motorcycle, bicycle, person and traffic light are the five worst-represented classes, thus the models should improve their performance in these classes when trained with uncertainty masks.

In Table \ref{tab:res_miou_exp1}, the class and mean IoU performance values for each backbone architecture are reported. The class IoU values of the five worst represented classes improved when using the instance-wise uncertainty (IU) masks to weight the pixels vs not using any weights. Most notably the rider class, which accounts for only $0.057\%$ of total pixels, saw improvements of up to $0.07$ of IoU. However, there are drops in IoU performance for the majority classes and there is no significant mIoU improvement. This is to be expected since the loss values of pixels of these classes will have lower weights than the others. These results suggest that using instance-wise uncertainty values can act as a regularization strategy for semantic segmentation deep learning models, as they can prevent overfitting to majority classes and greatly increase the performance on minority classes in extreme class imbalance situations.

\begin{table}[tb]
    \caption{The class IoU performance values for each different backbone architecture, according to the usage of uncertainty masks in their training. The "Regular" column and "IU" denote the regular training (no usage of instance uncertainty masks) and the instance-uncertainty weighted training (with instance-uncertainty masks). From left to right: ResNet50, ResNet101 and MobileNetV2.}
    \label{tab:res_miou_exp1}
    \begin{minipage}{0.32\textwidth}
        \centering
        \small
        \scalebox{0.7}{
        \begin{tabular}{|l|l|l|}
        \hline
            ~ & \textbf{Regular} & \textbf{IU} \\ \hline
            road & 0.952 \textpm 0.003 & 0.941 \textpm 0.002 \\ \hline
            sidewalk & 0.787 \textpm 0.006 & 0.755 \textpm 0.004\\ \hline
            building & 0.833 \textpm 0.003 & 0.827 \textpm 0.002\\ \hline
            wall & 0.51 \textpm 0.012 & 0.496 \textpm 0.008\\ \hline
            fence & 0.419 \textpm 0.012 & 0.4 \textpm 0.01 \\ \hline
            pole & 0.505 \textpm 0.008 & 0.504 \textpm 0.005\\ \hline
            traffic light & 0.633 \textpm 0.012 & 0.635 \textpm 0.006\\ \hline
            traffic sign & 0.528 \textpm 0.012 & 0.506 \textpm 0.007\\ \hline
            vegetation & 0.845 \textpm 0.003 & 0.841 \textpm 0.001 \\ \hline
            terrain & 0.487 \textpm 0.009 & 0.464 \textpm0.008\\ \hline
            sky & 0.947 \textpm 0.002 & 0.946 \textpm 0.001 \\ \hline
            person & 0.417 \textpm 0.018 & \textbf{0.454} \textpm 0.009 \\ \hline
            rider & 0.155 \textpm 0.063 & \textbf{0.204} \textpm 0.021 \\ \hline
            car & 0.825 \textpm 0.007 & 0.816 \textpm 0.005 \\ \hline
            truck & 0.411 \textpm 0.053 & 0.458 \textpm 0.025 \\ \hline
            bus & 0.744 \textpm 0.026 & 0.713 \textpm 0.035\\ \hline
            train & 0.847 \textpm 0.011 & 0.842 \textpm 0.009 \\ \hline
            motorcycle & 0.251 \textpm 0.032 & \textbf{0.291} \textpm 0.029\\ \hline
            bicycle & 0.359 \textpm 0.038 & \textbf{0.369} \textpm 0.031 \\ \hline
            \textbf{mIoU} & 0.604 & 0.604 \\ \hline
        \end{tabular}
        }
    \end{minipage}
    \hfill
    \begin{minipage}{0.32\textwidth}
        \centering
        \small
        \scalebox{0.7}{
        \begin{tabular}{|l|l|l|}
        \hline
            ~ & \textbf{Regular} & \textbf{IU} \\ \hline
            road & 0.954 \textpm 0.002 & 0.941 \textpm 0.002\\ \hline
            sidewalk & 0.794 \textpm 0.004 & 0.755 \textpm 0.004\\ \hline
            building & 0.836 \textpm 0.003 & 0.832 \textpm 0.002\\ \hline
            wall & 0.52 \textpm 0.008 & 0.509 \textpm 0.008 \\ \hline
            fence & 0.423 \textpm 0.012 & 0.422 \textpm 0.01 \\ \hline
            pole & 0.507 \textpm 0.006 & 0.509 \textpm 0.006 \\ \hline
            traffic light & 0.635 \textpm 0.008 & \textbf{0.647} \textpm 0.007\\ \hline
            traffic sign & 0.528 \textpm 0.009 & 0.533 \textpm 0.008 \\ \hline
            vegetation & 0.847 \textpm 0.003 & 0.846 \textpm 0.002\\ \hline
            terrain & 0.493 \textpm 0.01 & 0.476 \textpm 0.009 \\ \hline
            sky & 0.949 \textpm 0.002& 0.947 \textpm 0.002 \\ \hline
            person & 0.427 \textpm 0.021 & \textbf{0.492} \textpm 0.011 \\ \hline
            rider & 0.162 \textpm 0.053 & \textbf{0.231} \textpm 0.046 \\ \hline
            car & 0.828 \textpm 0.008 & 0.824 \textpm 0.005 \\ \hline
            truck & 0.486 \textpm 0.042 & 0.477 \textpm 0.02\\ \hline
            bus & 0.77 \textpm 0.027 & 0.695 \textpm 0.024\\ \hline
            train & 0.852 \textpm 0.013 & 0.831 \textpm 0.013\\ \hline
            motorcycle & 0.297 \textpm 0.038& \textbf{0.364} \textpm 0.02 \\ \hline
            bicycle & 0.346 \textpm 0.037 & \textbf{0.413} \textpm 0.038 \\ \hline
            \textbf{mIoU} & 0.614 & 0.619 \\ \hline
        \end{tabular}
        }
    \end{minipage}
    \hfill
    \begin{minipage}{0.32\textwidth}
        \centering
        \small
        \scalebox{0.7}{
        \begin{tabular}{|l|l|l|}
        \hline
            ~ & \textbf{Regular} & \textbf{IU} \\ \hline
            road & 0.944 \textpm 0.002 & 0.943 \textpm 0.002 \\ \hline
            sidewalk & 0.765 \textpm 0.005& 0.761 \textpm 0.006 \\ \hline
            building & 0.824 \textpm 0.002& 0.822 \textpm 0.003\\ \hline
            wall & 0.496 \textpm 0.009& 0.487 \textpm 0.008 \\ \hline
            fence & 0.401 \textpm 0.012& 0.39 \textpm 0.014\\ \hline
            pole & 0.491 \textpm 0.005& 0.495 \textpm 0.005\\ \hline
            traffic light & 0.587 \textpm 0.008 & \textbf{0.591} \textpm 0.01 \\ \hline
            traffic sign & 0.471 \textpm 0.01 & 0.478 \textpm 0.009 \\ \hline
            vegetation & 0.838 \textpm 0.002 & 0.836 \textpm 0.002 \\ \hline
            terrain & 0.469 \textpm 0.011 & 0.469 \textpm 0.008 \\ \hline
            sky & 0.946 \textpm 0.001 & 0.945 \textpm 0.001 \\ \hline
            person & 0.4 \textpm 0.018 & \textbf{0.416} \textpm 0.017\\ \hline
            rider & 0.201 \textpm 0.065& \textbf{0.221} \textpm 0.057 \\ \hline
            car & 0.798 \textpm 0.06 & 0.799 \textpm 0.005 \\ \hline
            truck & 0.372 \textpm 0.033& 0.386 \textpm 0.034 \\ \hline
            bus & 0.744 \textpm 0.023 & 0.731 \textpm 0.032 \\ \hline
            train & 0.822 \textpm 0.013 & 0.815 \textpm 0.015 \\ \hline
            motorcycle & 0.241 \textpm 0.037 & \textbf{0.264} \textpm 0.033 \\ \hline
            bicycle & 0.383 \textpm 0.031 & \textbf{0.397} \textpm 0.037 \\ \hline
            \textbf{mIoU} & 0.59 & 0.592 \\ \hline
        \end{tabular}
        }
    \end{minipage}
\end{table}

In this experiment, the usage of the inverse of class proportions as weights in the loss function hindered the IoU performance of the models. When compared to the usage of these weights, the gains in performance of the loss weighting with instance-wise uncertainty masks are even larger. In Table \ref{tab:res_miou_classprops}, the five worst-represented class IoU performance values for five different DeepLabV3+ models with the MobileNetV2 backbones trained with the inverse of class proportion as weights are reported. It should be noted that even though the weights for minority classes are much larger than the ones for majority classes, the models still can not improve their performance. These results further strengthen the idea of using uncertainty for improving performance on imbalanced datasets.

\begin{table}[tb]
    \caption{The class IoU performance values for the five worst-represented classes of five different DeepLabV3+ models with the MobileNetV2 backbone and the inverse of class proportions as loss weights.}
    \label{tab:res_miou_classprops}
    \centering
     \begin{tabular}{|l|l|}
       
        \hline
            ~ & \textbf{Class Proportions} \\ \hline
            traffic light & 0.557 \textpm 0.008 \\ \hline
            person & 0.398 \textpm 0.013 \\ \hline
            rider & 0.167 \textpm 0.019 \\ \hline
            motorcycle & 0.261 \textpm 0.023 \\ \hline
            bicycle & 0.341 \textpm 0.013 \\ \hline
            \textbf{mIoU} & 0.575 \\ \hline
    \end{tabular}
\end{table}

In Table \ref{tab:res_pavpu}, the PAvPU values for each backbone architeture are reported. The average pixel uncertainty of the train set was used as the certainty threshold for each patch, as is usual when using PAvPU. Weighting training samples with instance-wise uncertainty does not seem to improve the PAvPU of the models. This can be a result of the small performance drop in majority classes when using the uncertainty masks, leading the models to estimate larger uncertainties in their predictions for these pixels. This is in turn offset by the smaller estimated uncertainties for pixels of minority classes, ending with no overall PAvPU differences.

\begin{table}[tb]
    \caption{The PAvPU performance values for the DeepLabV3+ models with different backbones.}
    \label{tab:res_pavpu}
    \centering
     \begin{tabular}{|l|l|l|}
       
        \hline
            ~ & \textbf{Regular} & \textbf{IU} \\ \hline
            MobileNet & 0.832 \textpm 0.008 & 0.83 \textpm 0.007 \\ \hline
            ResNet50 & 0.849 \textpm 0.01 &  0.846 \textpm 0.004 \\ \hline
            ResNet101 & 0.852 \textpm 0.008 &  0.85 \textpm 0.004 \\ \hline
    \end{tabular}
\end{table}

The uncertainty masks can also be computed during the training process if the models are trained in parallel and combined in an ensemble. This corresponds to the second proposed method. Five different ensembles of three models (with different backbone architectures: ResNet50, ResNet101 and MobileNetV2) were trained by computing instance-wise uncertainty masks during training and their performance was compared to five different ensembles of models trained with no uncertainty masks. Their performance on the five worst represented classes is reported in Table \ref{tab:res_miou_end2end}. 

\begin{table}[tb]
    \caption{The class IoU performance values for the five worst-represented classes of five different ensembles of DeepLabV3+ models with the three backbones (ResNet50, ResNet101 and MobileNetV2). The IU masks are computed during training.}
    \label{tab:res_miou_end2end}
    \centering
     \begin{tabular}{|l|l|l|}
       
        \hline
            ~ & \textbf{Regular} & \textbf{Training IU} \\ \hline
            traffic light & 0.654 \textpm 0.002 & 0.664 \textpm 0.002\\ \hline
            person & 0.442 \textpm 0.009& \textbf{0.5} \textpm 0.012\\ \hline
            rider & 0.244 \textpm 0.035& 0.256 \textpm 0.029\\ \hline
            motorcycle & 0.265 \textpm 0.004 & \textbf{0.358} \textpm 0.02 \\ \hline
            bicycle & 0.417 \textpm 0.014 & \textbf{0.4417} \textpm 0.02 \\ \hline
            \textbf{mIoU} & 0.635 & 0.641\\ \hline
    \end{tabular}
\end{table}

In theory, computing the instance-wise uncertainty masks during training may lead to convergence issues due to poor uncertainty estimates by the untrained models, but this was not found to be the case since it actually improved model performance on minority classes. The reason for this may be that ensembles are usually better calibrated than individual models \cite{ovadia2019trustmodelsuncertaintyevaluating} and consequently predict accurate uncertainty estimates even when using models with weak performance. This suggests that instance-wise uncertainty masks are good loss weights regardless of the performance of the models that compute them.

\subsection{Improving model performance in domain-shift using instance-wise uncertainty}

To understand whether the proposed uncertainty masks improve the model's ability to generalize beyond its training data and its robustness to domain-shift, all models trained in the previous experiment were tested on the CityScapes validation set. Since CityScapes and ACDC are class compatible and their main difference is the presence of adverse weather conditions in ACDC, considering models trained on the ACDC dataset and evaluating them on CityScapes represents a very significant domain-shift from adverse weather to normal weather conditions. The class IoU values on the CityScapes validation set obtained by the 180 DeepLabV3+ models with different backbones trained on ACDC are reported in Table \ref{tab:res_miou_exp2}.

\begin{table}[tb]
    \caption{The class IoU performance values on the CityScapes validation set for each different backbone architecture, according to the usage of uncertainty masks in their training on ACDC. The "Regular" column and "IU" denote the regular training (no usage of instance uncertainty masks) and the instance-uncertainty weighted training (with instance-uncertainty masks). From left to right: ResNet50, ResNet101 and MobileNetV2.}
    \label{tab:res_miou_exp2}
    \begin{minipage}{0.32\textwidth}
        \centering
        \small
        \scalebox{0.7}{
        \begin{tabular}{|l|l|l|}
        \hline
            ~ & \textbf{Regular} & \textbf{IU} \\ \hline
            road & 0.788 \textpm 0.081 & \textbf{0.883} \textpm 0.018 \\ \hline
            sidewalk & 0.43 \textpm 0.053 & 0.478 \textpm 0.023 \\ \hline
            building & 0.735 \textpm 0.034 & \textbf{0.798} \textpm 0.009 \\ \hline
            wall & 0.122 \textpm 0.027 & \textbf{0.238} \textpm 0.022 \\ \hline
            fence & 0.183 \textpm  0.022 & 0.21 \textpm 0.014 \\ \hline
            pole & 0.265 \textpm 0.021 & 0.293 \textpm 0.01 \\ \hline
            traffic light & 0.211 \textpm 0.02 & \textbf{0.262} \textpm 0.014 \\ \hline
            traffic sign & 0.416 \textpm 0.018 & 0.422 \textpm 0.012 \\ \hline
            vegetation & 0.811 \textpm 0.016 & 0.824 \textpm 0.007 \\ \hline
            terrain & 0.081 \textpm 0.041 & 0.162 \textpm 0.024 \\ \hline
            sky & 0.745 \textpm 0.039 & 0.806 \textpm 0.012 \\ \hline
            person & 0.279 \textpm 0.036 & \textbf{0.452} \textpm 0.022 \\ \hline
            rider & 0.052 \textpm 0.018 & \textbf{0.128} \textpm 0.027 \\ \hline
            car & 0.792 \textpm 0.051 & 0.826 \textpm 0.012 \\ \hline
            truck & 0.159 \textpm 0.037 & \textbf{0.31} \textpm 0.022 \\ \hline
            bus & 0.251 \textpm 0.061 & \textbf{0.349} \textpm 0.038\\ \hline
            train & 0.049 \textpm 0.024 & 0.127 \textpm 0.048 \\ \hline
            motorcycle & 0.08 \textpm 0.021 & 0.115 \textpm 0.035\\ \hline
            bicycle & 0.42 \textpm 0.024 & \textbf{0.5} \textpm 0.01 \\ \hline
            \textbf{mIoU} & 0.361 & \textbf{0.431} \\ \hline
        \end{tabular}
        }
    \end{minipage}
    \hfill
    \begin{minipage}{0.32\textwidth}
        \centering
        \small
        \scalebox{0.7}{
        \begin{tabular}{|l|l|l|}
        \hline
            ~ & \textbf{Regular} & \textbf{IU} \\ \hline
            road & 0.848 \textpm 0.047 & \textbf{0.9} \textpm 0.018\\ \hline
            sidewalk & 0.468 \textpm 0.037 & \textbf{0.56} \textpm 0.025 \\ \hline
            building &  0.755 \textpm 0.028 & \textbf{0.816} \textpm 0.008 \\ \hline
            wall & 0.108 \textpm 0.027 & \textbf{0.214} \textpm 0.023 \\ \hline
            fence & 0.165 \textpm 0.015 & \textbf{0.227} \textpm 0.013 \\ \hline
            pole & 0.268 \textpm 0.014 & \textbf{0.314} \textpm 0.01 \\ \hline
            traffic light & 0.209 \textpm 0.016 & \textbf{0.309} \textpm 0.012 \\ \hline
            traffic sign & 0.41 \textpm 0.012 & 0.466 \textpm 0.013 \\ \hline
            vegetation & 0.815 \textpm 0.018 & 0.828 \textpm 0.01 \\ \hline
            terrain & 0.146 \textpm 0.066 & 0.258 \textpm 0.04 \\ \hline
            sky & 0.776 \textpm 0.033 & 0.818 \textpm 0.011 \\ \hline
            person & 0.253 \textpm 0.039 & \textbf{0.517} \textpm 0.015 \\ \hline
            rider & 0.068 \textpm 0.024 &  \textbf{0.203} \textpm 0.022 \\ \hline
            car & 0.8 \textpm 0.026 & 0.86 \textpm 0.003 \\ \hline
            truck & 0.185 \textpm 0.033 & \textbf{0.403} \textpm 0.028\\ \hline
            bus & 0.216 \textpm 0.05 & \textbf{0.382} \textpm 0.03 \\ \hline
            train & 0.065 \textpm 0.027 & 0.169 \textpm 0.056\\ \hline
            motorcycle & 0.064 \textpm 0.024 & \textbf{0.152} \textpm 0.034 \\ \hline
            bicycle & 0.384 \textpm 0.043 & \textbf{0.543} \textpm 0.011 \\ \hline
            \textbf{mIoU} & 0.369 & \textbf{0.47} \\ \hline
        \end{tabular}
        }
    \end{minipage}
    \hfill
    \begin{minipage}{0.32\textwidth}
        \centering
        \small
        \scalebox{0.7}{
        \begin{tabular}{|l|l|l|}
        \hline
            ~ & \textbf{Regular} & \textbf{IU} \\ \hline
            road &  0.802 \textpm 0.053 & 0.774 \textpm 0.064 \\ \hline
            sidewalk & 0.461 \textpm 0.029 & 0.445 \textpm 0.032 \\ \hline
            building & 0.774 \textpm 0.015 & 0.772 \textpm 0.023\\ \hline
            wall & 0.164 \textpm 0.027 & 0.149 \textpm 0.024 \\ \hline
            fence & 0.194 \textpm 0.029 & 0.19 \textpm 0.017\\ \hline
            pole & 0.28 \textpm 0.011& 0.279 \textpm 0.015\\ \hline
            traffic light & 0.188 \textpm 0.013 & 0.2 \textpm 0.017 \\ \hline
            traffic sign & 0.398 \textpm 0.011 & 0.402 \textpm 0.009 \\ \hline
            vegetation & 0.817 \textpm 0.014 & 0.812 \textpm 0.018 \\ \hline
            terrain & 0.099 \textpm 0.037 & 0.082 \textpm 0.029 \\ \hline
            sky & 0.752 \textpm 0.034 & 0.752 \textpm 0.026 \\ \hline
            person & 0.34 \textpm 0.031 & 0.341 \textpm 0.03\\ \hline
            rider & 0.112 \textpm 0.028 & 0.13 \textpm 0.035 \\ \hline
            car & 0.811 \textpm 0.013 & 0.801 \textpm 0.019 \\ \hline
            truck & 0.195 \textpm 0.039 & 0.192 \textpm 0.045 \\ \hline
            bus & 0.217 \textpm 0.051 & 0.217 \textpm 0.045 \\ \hline
            train & 0.099 \textpm 0.035 & 0.096 \textpm 0.041 \\ \hline
            motorcycle & 0.076 \textpm 0.028 & 0.08 \textpm 0.024 \\ \hline
            bicycle & 0.41 \textpm 0.024 & 0.435 \textpm 0.025 \\ \hline
            \textbf{mIoU} & 0.378 & 0.376 \\ \hline
        \end{tabular}
        }
    \end{minipage}
\end{table}

The mIoU values improve 0.07 and 0.1 when using instance-wise uncertainty masks for the models with the ResNet50 and ResNet101 backbones, respectively. Additionally to huge gains in IoU performance on the five worst represented classes, models with ResNet backbones also perform better on the majority classes when using uncertainty masks. Models with the MobileNetV2 backbone trained with uncertainty masks do not present much difference in IoU values to training with no uncertainty masks. The MobileNetV2 architecture is around 4.5 million parameters \cite{sandler2019mobilenetv2invertedresidualslinear}, which is much smaller than the ResNet50 and ResNet101 size. Due to this discrepancy in model size, the fact that using instance-wise uncertainty masks while training did not improve performance for the models with the MobileNetV2 backbone suggests that these uncertainties act as a regularization technique that decreases overfitting and improves generalization for more complex models.

%mpavpu on val set
%mpavpu on cityscapes test set
%iou on cityscapes test set
%ensemble training

\section{Conclusion}

In this work, a novel training approach for semantic segmentation that levarages instance-wise uncertainty to circumvent class imbalance was presented. Compared to currently available methods in the literature, the generation of uncertainty masks is done by computing the mean uncertainty of relevant neighborhood pixels, which corresponds to the mean uncertainty in each instance. By training as few as three models, an ensemble can be formed and used to compute an instance-wise uncertainty mask. These masks are then used to weight the cross entropy loss function, giving larger weights to pixels of class instances with higher uncertainty and vice-versa. The instance-wise uncertainty masks can be computed either during the training of an ensemble or by leveraging previously trained models. Although there is some computational overhead, models using uncertainty masks to weight their loss function were found to significantly improve their performance on minority classes, crucial for semantic segmentation due to class imbalance being very common, when compared to the usage of no loss weights or the usage of the inverse of class proportions. Instance-wise uncertainty also drastically improved the generalization and reduced the overfitting of some models, which leads to safer real world deployment. Reducing the computational overhead of computing instance-wise uncertainty masks and finding other ways to integrate them in the training process represents an interesting path for future work in improving semantic segmentation performance.

\bibliographystyle{splncs04}
\bibliography{main.bib}

\begin{thebibliography}{10}
\providecommand{\url}[1]{\texttt{#1}}
\providecommand{\urlprefix}{URL }
\providecommand{\doi}[1]{https://doi.org/#1}

\bibitem{ashukha2021pitfalls}
Ashukha, A., Lyzhov, A., Molchanov, D., Vetrov, D.: Pitfalls of in-domain uncertainty estimation and ensembling in deep learning (2021), \url{https://arxiv.org/abs/2002.06470}, arXiv preprint

\bibitem{pmlr-v37-blundell15}
Blundell, C., Cornebise, J., Kavukcuoglu, K., Wierstra, D.: Weight uncertainty in neural network. In: Bach, F., Blei, D. (eds.) Proceedings of the 32nd International Conference on Machine Learning. Proceedings of Machine Learning Research, vol.~37, pp. 1613--1622. PMLR, Lille, France (07--09 Jul 2015), \url{https://proceedings.mlr.press/v37/blundell15.html}

\bibitem{BRESSAN2022102690}
Bressan, P.O., Junior, J.M., {Correa Martins}, J.A., {de Melo}, M.J., Gonçalves, D.N., Freitas, D.M., {Marques Ramos}, A.P., {Garcia Furuya}, M.T., Osco, L.P., {de Andrade Silva}, J., Luo, Z., Garcia, R.C., Ma, L., Li, J., Gonçalves, W.N.: Semantic segmentation with labeling uncertainty and class imbalance applied to vegetation mapping. International Journal of Applied Earth Observation and Geoinformation  \textbf{108},  102690 (2022). \doi{https://doi.org/10.1016/j.jag.2022.102690}, \url{https://www.sciencedirect.com/science/article/pii/S0303243422000162}

\bibitem{Brier1950VERIFICATIONOF}
Brier, G.W.: Verification of forecasts expressed in terms of probability. Monthly Weather Review  \textbf{78}, ~1--3 (1950)

\bibitem{caplin2022calibratingclassweightsmodeling}
Caplin, A., Martin, D., Marx, P.: Calibrating for class weights by modeling machine learning (2022), \url{https://arxiv.org/abs/2205.04613}, arXiv preprint

\bibitem{Carvalho_2020_CVPR}
Carvalho, E.D.C., Clark, R., Nicastro, A., Kelly, P.H.J.: Scalable uncertainty for computer vision with functional variational inference. In: Proceedings of the IEEE/CVF Conference on Computer Vision and Pattern Recognition (CVPR) (June 2020)

\bibitem{DBLP:journals/corr/abs-1802-02611}
Chen, L., Zhu, Y., Papandreou, G., Schroff, F., Adam, H.: Encoder-decoder with atrous separable convolution for semantic image segmentation. CoRR  \textbf{abs/1802.02611} (2018), \url{http://arxiv.org/abs/1802.02611}

\bibitem{DBLP:journals/corr/CordtsORREBFRS16}
Cordts, M., Omran, M., Ramos, S., Rehfeld, T., Enzweiler, M., Benenson, R., Franke, U., Roth, S., Schiele, B.: The cityscapes dataset for semantic urban scene understanding. CoRR  \textbf{abs/1604.01685} (2016), \url{http://arxiv.org/abs/1604.01685}

\bibitem{9533330}
Cygert, S., Wróblewski, B., Woźniak, K., Słowiński, R., Czyżewski, A.: Closer look at the uncertainty estimation in semantic segmentation under distributional shift. In: 2021 International Joint Conference on Neural Networks (IJCNN). pp.~1--8 (2021). \doi{10.1109/IJCNN52387.2021.9533330}

\bibitem{9956231}
Egele, R., Maulik, R., Raghavan, K., Lusch, B., Guyon, I., Balaprakash, P.: Autodeuq: Automated deep ensemble with uncertainty quantification. In: 2022 26th International Conference on Pattern Recognition (ICPR). pp. 1908--1914 (2022). \doi{10.1109/ICPR56361.2022.9956231}

\bibitem{18631}
Feiwei, Q., Xiyue, S., Yong, P., Yanli, S., Wenqiang, Y., Zhongping, J., Jing, B.: A real-time semantic segmentation approach for autonomous driving scenes. Journal of Computer-Aided Design \& Computer Graphics  \textbf{33}(7),  1026--1037 (2021). \doi{10.3724/SP.J.1089.2021.18631}, \url{https://www.jcad.cn/en/article/doi/10.3724/SP.J.1089.2021.18631}

\bibitem{DBLP:journals/corr/abs-2110-04286}
Folgoc, L.L., Baltatzis, V., Desai, S., Devaraj, A., Ellis, S., Manzanera, O.E.M., Nair, A., Qiu, H., Schnabel, J.A., Glocker, B.: Is {MC} dropout bayesian? CoRR  \textbf{abs/2110.04286} (2021), \url{https://arxiv.org/abs/2110.04286}

\bibitem{pmlr-v48-gal16}
Gal, Y., Ghahramani, Z.: Dropout as a bayesian approximation: Representing model uncertainty in deep learning. In: Balcan, M.F., Weinberger, K.Q. (eds.) Proceedings of The 33rd International Conference on Machine Learning. Proceedings of Machine Learning Research, vol.~48, pp. 1050--1059. PMLR, New York, New York, USA (20--22 Jun 2016), \url{https://proceedings.mlr.press/v48/gal16.html}

\bibitem{10.1007/s10462-023-10562-9}
Gawlikowski, J., Tassi, C.R.N., Ali, M., Lee, J., Humt, M., Feng, J., Kruspe, A., Triebel, R., Jung, P., Roscher, R., Shahzad, M., Yang, W., Bamler, R., Zhu, X.X.: A survey of uncertainty in deep neural networks. Artif. Intell. Rev.  \textbf{56}(Suppl 1),  1513–1589 (jul 2023). \doi{10.1007/s10462-023-10562-9}, \url{https://doi.org/10.1007/s10462-023-10562-9}

\bibitem{DBLP:journals/corr/GuoPSW17}
Guo, C., Pleiss, G., Sun, Y., Weinberger, K.Q.: On calibration of modern neural networks. CoRR  \textbf{abs/1706.04599} (2017), \url{http://arxiv.org/abs/1706.04599}

\bibitem{resnet50}
He, K., Zhang, X., Ren, S., Sun, J.: Deep residual learning for image recognition. CoRR  \textbf{abs/1512.03385} (2015), \url{http://arxiv.org/abs/1512.03385}

\bibitem{DBLP:journals/corr/HowardZCKWWAA17}
Howard, A.G., Zhu, M., Chen, B., Kalenichenko, D., Wang, W., Weyand, T., Andreetto, M., Adam, H.: Mobilenets: Efficient convolutional neural networks for mobile vision applications. CoRR  \textbf{abs/1704.04861} (2017), \url{http://arxiv.org/abs/1704.04861}

\bibitem{kahl2024values}
Kahl, K.C., Lüth, C.T., Zenk, M., Maier-Hein, K., Jaeger, P.F.: Values: A framework for systematic validation of uncertainty estimation in semantic segmentation (2024), \url{https://arxiv.org/abs/2401.08501}, arXiv preprint

\bibitem{Kar2021}
Kar, M.K., Nath, M.K., Neog, D.R.: A review on progress in semantic image segmentation and its application to medical images. SN Computer Science  \textbf{2}(5), ~397 (2021). \doi{10.1007/s42979-021-00784-5}, \url{https://doi.org/10.1007/s42979-021-00784-5}

\bibitem{KIUREGHIAN2009105}
Kiureghian, A.D., Ditlevsen, O.: Aleatory or epistemic? does it matter? Structural Safety  \textbf{31}(2),  105--112 (2009). \doi{https://doi.org/10.1016/j.strusafe.2008.06.020}, \url{https://www.sciencedirect.com/science/article/pii/S0167473008000556}, risk Acceptance and Risk Communication

\bibitem{NIPS2017_9ef2ed4b}
Lakshminarayanan, B., Pritzel, A., Blundell, C.: Simple and scalable predictive uncertainty estimation using deep ensembles. In: Guyon, I., Luxburg, U.V., Bengio, S., Wallach, H., Fergus, R., Vishwanathan, S., Garnett, R. (eds.) Advances in Neural Information Processing Systems. vol.~30. Curran Associates, Inc. (2017), \url{https://proceedings.neurips.cc/paper_files/paper/2017/file/9ef2ed4b7fd2c810847ffa5fa85bce38-Paper.pdf}

\bibitem{landgraf2023uceuncertaintyawarecrossentropysemantic}
Landgraf, S., Hillemann, M., Wursthorn, K., Ulrich, M.: U-ce: Uncertainty-aware cross-entropy for semantic segmentation (2023), \url{https://arxiv.org/abs/2307.09947}, arXiv preprint

\bibitem{malinin2019uncertainty}
Malinin, A.: Uncertainty Estimation in Deep Learning with application to Spoken Language Assessment. Ph.D. thesis, Department Of Engineering, University Of Cambridge (2019). \doi{10.17863/CAM.45912}, \url{https://doi.org/10.17863/CAM.45912}

\bibitem{Mobiny2021}
Mobiny, A., Yuan, P., Moulik, S.K., Garg, N., Wu, C.C., Nguyen, H.V.: Dropconnect is effective in modeling uncertainty of bayesian deep networks. Scientific Reports  \textbf{11}(1), ~5458 (March 9 2021). \doi{10.1038/s41598-021-84854-x}, \url{https://doi.org/10.1038/s41598-021-84854-x}

\bibitem{9913352}
Muhammad, K., Hussain, T., Ullah, H., Ser, J.D., Rezaei, M., Kumar, N., Hijji, M., Bellavista, P., de~Albuquerque, V.H.C.: Vision-based semantic segmentation in scene understanding for autonomous driving: Recent achievements, challenges, and outlooks. IEEE Transactions on Intelligent Transportation Systems  \textbf{23}(12),  22694--22715 (2022). \doi{10.1109/TITS.2022.3207665}

\bibitem{DBLP:journals/corr/abs-1811-12709}
Mukhoti, J., Gal, Y.: Evaluating bayesian deep learning methods for semantic segmentation. CoRR  \textbf{abs/1811.12709} (2018), \url{http://arxiv.org/abs/1811.12709}

\bibitem{DBLP:journals/corr/abs-2102-11582}
Mukhoti, J., Kirsch, A., van Amersfoort, J., Torr, P.H.S., Gal, Y.: Deterministic neural networks with appropriate inductive biases capture epistemic and aleatoric uncertainty. CoRR  \textbf{abs/2102.11582} (2021), \url{https://arxiv.org/abs/2102.11582}

\bibitem{Naeini2015ObtainingWC}
Naeini, M.P., Cooper, G.F., Hauskrecht, M.: Obtaining well calibrated probabilities using bayesian binning. Proc Conf AAAI Artif Intell pp. 2901--2907 (2015)

\bibitem{Opper2009}
Opper, M., Archambeau, C.: The variational gaussian approximation revisited. Neural Computation  \textbf{21}(3),  786--792 (March 2009). \doi{10.1162/neco.2008.08-07-592}, department of Computer Science, Technical University Berlin, D-10587 Berlin, Germany. opperm@cs.tu-berlin.de

\bibitem{ovadia2019trustmodelsuncertaintyevaluating}
Ovadia, Y., Fertig, E., Ren, J., Nado, Z., Sculley, D., Nowozin, S., Dillon, J.V., Lakshminarayanan, B., Snoek, J.: Can you trust your model's uncertainty? evaluating predictive uncertainty under dataset shift (2019), \url{https://arxiv.org/abs/1906.02530}

\bibitem{postels2022practicalitydeterministicepistemicuncertainty}
Postels, J., Segu, M., Sun, T., Sieber, L., Gool, L.V., Yu, F., Tombari, F.: On the practicality of deterministic epistemic uncertainty (2022), \url{https://arxiv.org/abs/2107.00649}, arXiv preprint

\bibitem{DBLP:journals/corr/abs-2104-13395}
Sakaridis, C., Dai, D., Gool, L.V.: {ACDC:} the adverse conditions dataset with correspondences for semantic driving scene understanding. CoRR  \textbf{abs/2104.13395} (2021), \url{https://arxiv.org/abs/2104.13395}

\bibitem{sandler2019mobilenetv2invertedresidualslinear}
Sandler, M., Howard, A., Zhu, M., Zhmoginov, A., Chen, L.C.: Mobilenetv2: Inverted residuals and linear bottlenecks (2019), \url{https://arxiv.org/abs/1801.04381}

\bibitem{8317714}
Siam, M., Elkerdawy, S., Jagersand, M., Yogamani, S.: Deep semantic segmentation for automated driving: Taxonomy, roadmap and challenges. In: 2017 IEEE 20th International Conference on Intelligent Transportation Systems (ITSC). pp.~1--8 (2017). \doi{10.1109/ITSC.2017.8317714}

\bibitem{singh2022quantifyingmodeluncertaintysemantic}
Singh, R., Principe, J.C.: Quantifying model uncertainty for semantic segmentation using operators in the rkhs (2022), \url{https://arxiv.org/abs/2211.01999}, arXiv preprint

\bibitem{smith2018understandingmeasuresuncertaintyadversarial}
Smith, L., Gal, Y.: Understanding measures of uncertainty for adversarial example detection (2018), \url{https://arxiv.org/abs/1803.08533}, arXiv preprint

\bibitem{JMLR:v15:srivastava14a}
Srivastava, N., Hinton, G., Krizhevsky, A., Sutskever, I., Salakhutdinov, R.: Dropout: A simple way to prevent neural networks from overfitting. Journal of Machine Learning Research  \textbf{15}(56),  1929--1958 (2014), \url{http://jmlr.org/papers/v15/srivastava14a.html}

\bibitem{article}
V, B.: Biomedical image analysis using semantic segmentation. Journal of Innovative Image Processing  \textbf{1},  91--101 (12 2019). \doi{10.36548/jiip.2019.2.004}

\bibitem{fastdropouttraining}
Wang, S., Manning, C.: Fast dropout training. In: Dasgupta, S., McAllester, D. (eds.) Proceedings of the 30th International Conference on Machine Learning. Proceedings of Machine Learning Research, vol.~28, pp. 118--126. PMLR, Atlanta, Georgia, USA (17--19 Jun 2013), \url{https://proceedings.mlr.press/v28/wang13a.html}

\bibitem{9330594}
Weng, W., Zhu, X.: Inet: Convolutional networks for biomedical image segmentation. IEEE Access  \textbf{9},  16591--16603 (2021). \doi{10.1109/ACCESS.2021.3053408}

\bibitem{wimmer2023quantifying}
Wimmer, L., Sale, Y., Hofman, P., Bischl, B., Hüllermeier, E.: Quantifying aleatoric and epistemic uncertainty in machine learning: Are conditional entropy and mutual information appropriate measures? (2023), \url{https://arxiv.org/abs/2209.03302}, arXiv preprint

\end{thebibliography}
\end{document}